\documentclass[11pt,a4paper]{article}
\usepackage[hyperref]{eacl2021}
\usepackage{arabtex}
\usepackage{utf8}

\usepackage{latexsym}

\usepackage{microtype}

\usepackage{times}
\usepackage{url}
\usepackage{latexsym}
\usepackage{multirow}
\usepackage{makecell}
\usepackage{graphicx}
\usepackage{url}
\usepackage{paralist}
\usepackage{tabularx}
\usepackage{qtree}

\usepackage{subcaption}
\usepackage{wrapfig,lipsum,booktabs}

\usepackage{colortbl}
\usepackage{siunitx}
\usepackage{color,soul}

\usepackage{comment}

\usepackage{adjustbox}
\usepackage{appendix}
\usepackage{placeins}

\usepackage{tikz}
\def\checkmark{\tikz\fill[scale=0.4](0,.35) -- (.25,0) -- (1,.7) -- (.25,.15) -- cycle;} 

\definecolor{light-gray}{gray}{0.85}

\usepackage{arydshln}

\usepackage{cancel}
\usepackage{ulem,xpatch}

\xpatchcmd{\sout}
  {\bgroup}
  {\bgroup}
  {}{}

\newcommand{\eat}[1]{}

\usepackage{footnote}
\makesavenoteenv{tabular}
\makesavenoteenv{table}
\usepackage{changepage}

\usepackage{microtype}

\aclfinalcopy 

\newcolumntype{H}{>{\setbox0=\hbox\bgroup}c<{\egroup}@{}}

\title{NADI 2021:\\The Second Nuanced Arabic Dialect Identification Shared Task}

\author{Muhammad Abdul-Mageed, Chiyu Zhang, AbdelRahim Elmadany, \\\textbf{Houda Bouamor},$^\dagger$ \textbf{Nizar Habash}$^\ddagger$\\
The University of British Columbia, Vancouver, Canada\\
$^\dagger$Carnegie Mellon University in Qatar, Qatar\\
$^\ddagger$New York University Abu Dhabi, UAE\\
  {\tt \{muhammad.mageed, a.elmadany\}@ubc.ca} \tt ~~~~ chiyuzh@mail.ubc.ca\\
  {\tt ~~ hbouamor@cmu.edu ~~~~ nizar.habash@nyu.edu}\\
  }
\date{}

\begin{document}

\setarab 
\maketitle
\setcode{utf8}
\centerline{\large\bf Abstract}%
\vspace{0.25ex}

\begin{adjustwidth}{6pt}{6pt}
 \noindent 
We present the findings and results of the Second Nuanced Arabic Dialect Identification Shared Task (NADI 2021). 
This Shared Task includes four subtasks: country-level Modern Standard Arabic (MSA) identification (Subtask 1.1), country-level dialect identification (Subtask 1.2), province-level MSA identification (Subtask 2.1), and province-level sub-dialect identification (Subtask 2.2). The shared task dataset covers a total of 100 provinces from 21 Arab countries, collected from the Twitter domain.  
A total of 53 teams from 23 countries registered to participate in the tasks, thus reflecting the interest of the community in this area. We received 16 submissions for Subtask 1.1 from five teams, 27 submissions for Subtask 1.2 from eight teams, 12 submissions for Subtask 2.1 from four teams, and 13 Submissions for subtask 2.2 from four teams.
\end{adjustwidth}

\section{Introduction}\label{sec:intro}
Arabic is the native tongue of $\sim 400$ million people living the Arab world, a vast geographical region across Africa and Asia. Far from a single monolithic language, Arabic has a wide number of varieties. In general, Arabic could be classified into three main categories: (1) Classical Arabic, the language of the Qur'an and early literature; (2) Modern Standard Arabic (MSA), which is usually used in education and formal and pan-Arab media; and (3) dialectal Arabic (DA), a collection of geo-politically defined variants. Modern day Arabic is usually referred to as \textit{diglossic} with a so-called `High' variety used in formal settings (MSA), and a `Low' variety used in everyday communication (DA). DA, the presumably `Low' variety, is itself a host of variants. For the current work, we focus on geography as an axis of variation where people from various sub-regions, countries, or even provinces within the same country, may be using Arabic differently. 

\begin{figure}[t]
  \begin{center}
  \frame{\includegraphics[width=8cm,height=4.5cm]{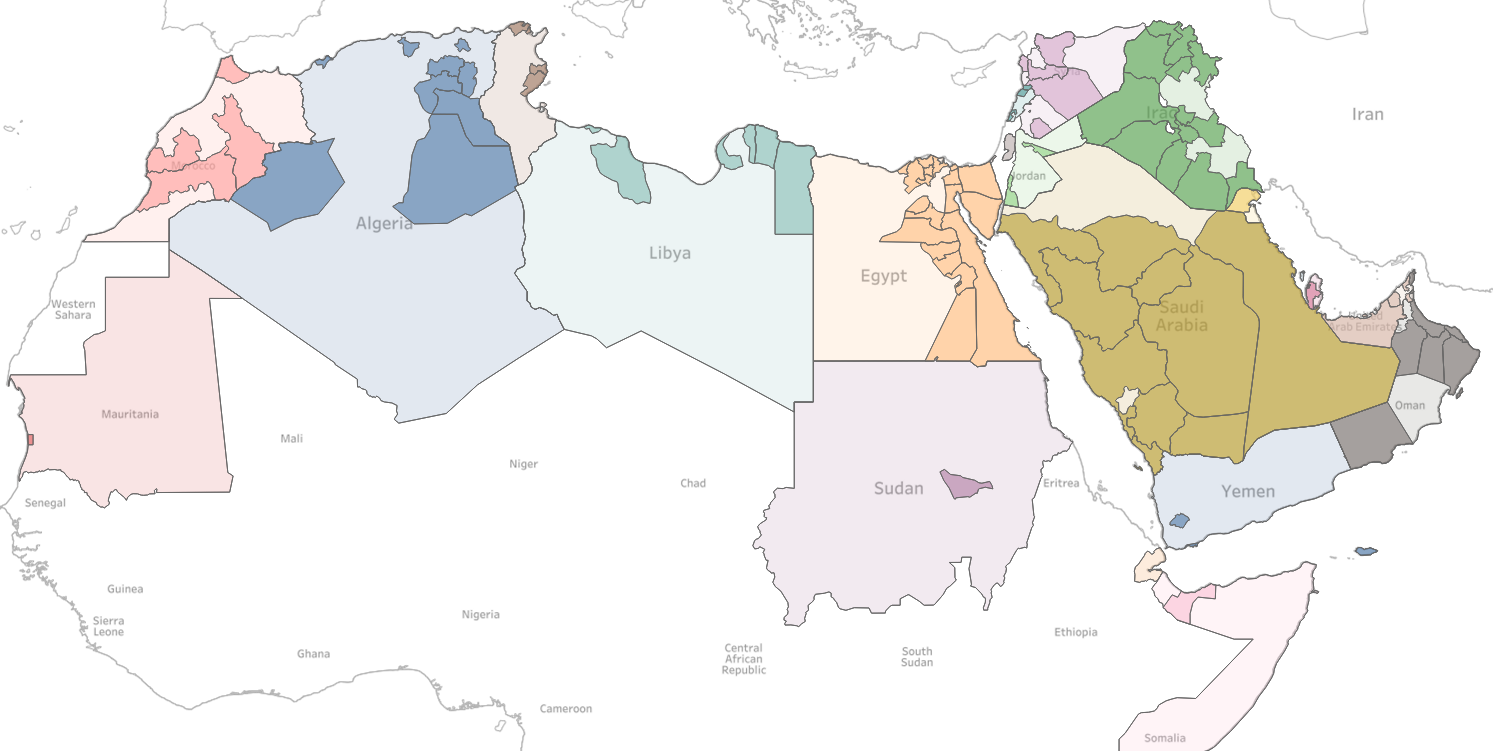}}
  \end{center}
\caption{A map of the Arab World showing the 21 countries and 100 provinces in the NADI 2021 datasets. Each country is coded in color different from neighboring countries. Provinces within each country are coded in a more intense version of the same color as the country.}
\label{fig:province_map}
\end{figure}

The Nuanced Arabic Dialect Identification (NADI) series of shared tasks aim at furthering the study and analysis of Arabic variants by providing resources and organizing classification competitions under standardized settings. The First Nuanced Arabic Dialect Identification (NADI 2020) Shared Task targeted 21 Arab countries and a total of 100 provinces across these countries. NADI 2020 consisted of two subtasks: \textit{country-level} dialect identification (Subtask~1) and  \textit{province-level} detection (Subtask~2). The two subtasks depended on Twitter data, making it the first shared task to target naturally-occurring fine-grained dialectal text at the sub-country level. The Second Nuanced Arabic Dialect Identification (NADI 2021) is similar to NADI 2020 in that it also targets the same 21 Arab countries and 100 corresponding provinces and is based on Twitter data. However, NADI 2021 has four subtasks, organized into country level and province level. For each classification level, we afford both MSA and DA datasets as Table~\ref{tab:tasks} shows.

\begin{table}[h]
\centering
\begin{tabular}{lll} \hline 
\textbf{Variety} & \textbf{Country}      & \textbf{Province}     \\ \hline 
MSA  & Subtask 1.1 & Subtask 2.1 \\
DA   & Subtask 1.2 & Subtask 2.2 \\ \hline 
\end{tabular}
\caption{NADI 2021 subtasks.}
\label{tab:tasks}
\end{table}

We provided participants with a new Twitter labeled dataset that we collected exclusively for the purpose of the shared task. 
The dataset is publicly available for research.\footnote{The dataset is accessible via our GitHub at: \url{https://github.com/UBC-NLP/nadi}.} A total of 53 teams registered for the shard task, of whom 8 unique teams ended up submitting their systems for scoring. We allowed a maximum of five submissions per team. We received 16 submissions for Subtask 1.1 from five teams, 27 submissions for Subtask 1.2 from eight teams, 12 submissions for Subtask 2.1 from four teams, and 13 Submissions for subtask 2.2 from four teams. We then received seven papers, all of which we accepted for publication.

This paper is organized as follows.  We provide a brief overview of the computational linguistic literature on Arabic dialects in Section~\ref{sec:rel}. 
We describe the two subtasks and dataset in Sections~\ref{sec:task} and Section~\ref{sec:data}, respectively. And finally, we introduce participating teams, shared task results, and a high-level description of submitted systems in Section~\ref{sec:teams_res}.

\section{Related Work}\label{sec:rel}
 As we explained in Section~\ref{sec:intro}, Arabic has three main categories: CA, MSA, and DA. While CA and MSA have been studied extensively~\cite{Harrell:1962:short,Cowell:1964:reference,badawi1973levels,Brustad:2000:syntax,Holes:2004:modern}, DA has received more attention only in recent years. 
 
 One major challenge with studying DA has been rarity of resources. For this reason, most pioneering DA works focused on creating resources, usually for only a small number of regions or countries~\cite{Gadalla:1997:callhome,diab2010colaba,al2012yadac,sadat2014automatic,Smaili:2014:building,Jarrar:2016:curras,Khalifa:2016:large,Al-Twairesh:2018:suar,el-haj-2020-habibi}. A number of works introducing multi-dialectal data sets and regional level detection models followed~\cite{zaidan2011arabic,Elfardy:2014:aida,Bouamor:2014:multidialectal,Meftouh:2015:machine}. 
 
 Arabic dialect identification work as further sparked by a series of shared tasks offered as part of the VarDial workshop. These shared tasks used speech broadcast transcriptions~\cite{malmasi2016discriminating}, and integrated acoustic features~\cite{zampieri2017findings} and phonetic features~\cite{zampieri2018language} extracted from raw audio.  \newcite{althobaiti2020automatic} is a recent survey of computational work on Arabic dialects.

The Multi Arabic Dialects Application and Resources (MADAR) project ~\cite{Bouamor:2018:madar} introduced finer-grained dialectal data and a lexicon. The MADAR data were used for dialect identification at the city level \cite{Salameh:2018:fine-grained,obeid-etal-2019-adida} of 25 Arab cities.  An issue with the MADAR data, in the context of DA identification, is that it was commissioned and not naturally occurring. Several larger datasets covering 10-21 countries were also introduced~\cite{Mubarak:2014:using,Abdul-Mageed:2018:you,Zaghouani:2018:araptweet}. These datasets come from the Twitter domain, and hence are naturally-occurring. 

Several works have also focused on socio-pragmatics meaning exploiting dialectal data. These include sentiment analysis ~\cite{Abdul-Mageed:2014:samar}, emotion~\cite{alhuzali2018enabling}, age and gender~\cite{abbes2020daict}, offensive language~\cite{mubarak2020overview}, and sarcasm~\cite{farha2020arabic}. Concurrent with our work,~\cite{mageed2020microdialect} also describe data and models at country, province, and city levels.

The first NADI shared task, NADI 2020~\cite{mageed:2020:nadi}, comprised two subtasks, one focusing on 21 Arab countries exploiting Twitter data, and another on 100 Arab provinces from the same 21 countries. As is explained in~\cite{mageed:2020:nadi}, the NADI 2020 datasets included a small amount of non-Arabic and also a mixture of MSA and DA. For NADI 2021, we continue to focus on 21 countries and 100 provinces. However, we breakdown the data into MSA and DA for a stronger signal. This also gives us the opportunity to study each of these two main categories independently. In other words, in addition to dialect and sub-dialect identification, it allows us to investigate the extent to which MSA itself can be teased apart at the country and province levels. Our hope is that NADI 2021 will support exploring variation in geographical regions that have not been studied before.

\section{Task Description}\label{sec:task}
\begin{table*}[t]
\centering
\small

\begin{tabular}{lcrrrrr|rrrrr}
\toprule

                                            &                                           & \multicolumn{5}{c|}{\textbf{MSA (Subtasks 1.1 \& 2.1)}}                                            & \multicolumn{5}{c}{\textbf{DA (Subtasks 1.2 \& 2.2)}}                                             \\ \cline{3-12} 
\multirow{-2}{*}{\textbf{Country}}          & \multirow{-2}{*}{\textbf{Provinces}} & \textbf{Train} & \textbf{DEV} & \textbf{TEST} & \textbf{Total} & \textbf{\%} & \textbf{Train} & \textbf{DEV} & \textbf{TEST} & \textbf{Total} & \textbf{\%} \\ \toprule
Algeria                & 9                                         & 1,899          & 427          & 439           & 2,765          & 8.92      & 1,809          & 430          & 391           & 2,630          & 8.48      \\ 
Bahrain                & 1                                         & 211            & 51           & 51            & 313            & 1.01      & 215            & 52           & 52            & 319            & 1.03      \\ 
Djibouti               & 1                                         & 211            & 52           & 51            & 314            & 1.01      & 215            & 27           & 7             & 249            & 0.80      \\ 
Egypt                  & 20                                        & 4,220          & 1,032        & 989           & 6,241          & 20.13     & 4,283          & 1,041        & 1,051         & 6,375          & 20.56     \\ 
Iraq                   & 13                                        & 2,719          & 671          & 652           & 4,042          & 13.04     & 2,729          & 664          & 664           & 4,057          & 13.09     \\ 
Jordan                 & 2                                         & 422            & 103          & 102           & 627            & 2.02      & 429            & 104          & 105           & 638            & 2.06      \\ 
Kuwait                 & 2                                         & 422            & 103          & 102           & 627            & 2.02      & 429            & 105          & 106           & 640            & 2.06      \\ 
Lebanon                & 3                                         & 633            & 155          & 141           & 929            & 3.00      & 644            & 157          & 120           & 921            & 2.97      \\ 
Libya                  & 6                                         & 1,266          & 310          & 307           & 1,883          & 6.07      & 1,286          & 314          & 316           & 1,916          & 6.18      \\ 
Mauritania             & 1                                         & 211            & 52           & 51            & 314            & 1.01      & 215            & 53           & 53            & 321            & 1.04      \\ 
Morocco                & 4                                         & 844            & 207          & 205           & 1,256          & 4.05      & 858            & 207          & 212           & 1,277          & 4.12      \\ 
Oman                   & 7                                         & 1,477          & 341          & 357           & 2,175          & 7.02      & 1,501          & 355          & 371           & 2,227          & 7.18      \\ 
Palestine              & 2                                         & 422            & 102          & 102           & 626            & 2.02      & 428            & 104          & 105           & 637            & 2.05      \\ 
Qatar                  & 1                                         & 211            & 52           & 51            & 314            & 1.01      & 215            & 52           & 53            & 320            & 1.03      \\ 
KSA          & 10                                        & 2,110          & 510          & 510           & 3,130          & 10.10     & 2,140          & 520          & 522           & 3,182          & 10.26     \\ 
Somalia                & 2                                         & 346            & 63           & 102           & 511            & 1.65      & 172            & 49           & 55            & 276            & 0.89      \\ 
Sudan                  & 1                                         & 211            & 48           & 51            & 310            & 1.00      & 215            & 53           & 53            & 321            & 1.04      \\ 
Syria                  & 6                                         & 1,266          & 309          & 306           & 1,881          & 6.07      & 1,287          & 278          & 288           & 1,853          & 5.98      \\ 
Tunisia                & 4                                         & 844            & 170          & 176           & 1,190          & 3.84      & 859            & 173          & 212           & 1,244          & 4.01      \\ 
UAE & 3                                         & 633            & 154          & 153           & 940            & 3.03      & 642            & 157          & 158           & 957            & 3.09      \\ 
Yemen                  & 2                                         & 422            & 88           & 102           & 612            & 1.97      & 429            & 105          & 106           & 640            & 2.06      \\ \toprule
\textbf{Total}                  & \textbf{100}                                       & \textbf{21,000}         & \textbf{5,000}        & \textbf{5,000}         & \textbf{31,000}         & \textbf{100}           & \textbf{21,000}         & \textbf{5,000}        & \textbf{5,000}         & \textbf{31,000}          & \textbf{100 }          \\
\toprule
\end{tabular}%
\caption{Distribution of classes and data splits over our MSA and DA datasets for the four subtasks.}.\label{tab:country}
\end{table*}

The NADI shared task consists of four subtasks, comprising two levels of classification--country and province. Each level of classification is carried out for both MSA and DA. We explain the different subtasks across each classification level next. 

\subsection{Country-level Classification}

\begin{itemize}
    \item \textbf{Subtask~1.1: Country-level MSA.}  The goal of Subtask~1.1 is to identify country level MSA from short written sentences (tweets). NADI 2021 Subtask~1.1 is novel since no previous works focused on teasing apart MSA by country of origin.
    \item \noindent\textbf{Subtask~1.2: Country-level DA.} Subtask~1.2 is similar to Subtask~1.1, but focuses on identifying country level \textit{dialect} from tweets. Subtask~1.2 is similar to previous works that have also taken country as their target  \cite{Mubarak:2014:using,Abdul-Mageed:2018:you,Zaghouani:2018:araptweet,bouamor2019madar,mageed:2020:nadi}.
\end{itemize}
 
We provided labeled data to NADI 2021 participants with specific training (TRAIN) and development (DEV) splits. Each of the 21 labels corresponding to the 21 countries is represented in both TRAIN and DEV. Teams could score their models through an online system (codalab) on the DEV set before the deadline. We released our TEST set of unlabeled tweets shortly before the system submission deadline. We then invited participants to submit their predictions to the online scoring system housing the gold TEST set labels. Table~\ref{tab:country} shows the distribution of the TRAIN, DEV, and TEST splits across the 21 countries.

\subsection{Province-level Classification}


\begin{itemize}
    \item \textbf{Subtask~2.1: Province-level MSA.} The goal of Subtask~2.1 is to identify the specific state or province (henceforth, {\it province}) from which an MSA tweet was posted. There are 100 province labels in the data, and provinces are unequally distributed among the list of 21 countries.
    \item \textbf{Subtask~2.2: Province-level DA.}
    Again, Subtask~2.2 is similar to Subtask~2.1, but the goal is identifying the province from which a \textit{dialectal} tweet was posted. 
\end{itemize}

 While the MADAR~ shared task~\cite{bouamor2019madar} involved prediction of a small set of cities, NADI 2020 was the first to propose automatic dialect identification at geographical regions as small as provinces. Concurrent with NADI 2020,~\cite{mageed2020microdialect} introduced the concept of \textit{microdialects}, and proposed models for identifying language varieties defined at both province and city levels. NADI 2021 follows these works, but has one novel aspect: We introduce province-level identification for MSA and DA independently (i.e., each variety is handled in a separate subtask). While province-level sub-dialect identification may be challenging, we hypothesize province-level MSA might be even more difficult. However, we were curious to what extent, if possible at all, a machine would be successful in teasing apart MSA data at the province-level.  
 
In addition, similar to NADI 2020, we acknowledge that province-level classification is somewhat related to geolocation prediction exploiting Twitter data. However, we emphasize that geolocation prediction is performed at the level of \textit{users}, rather than tweets. This makes our subtasks different from geolocation work. Another difference lies in the way we collect our data as we will explain in Section~\ref{sec:data}. Tables~\ref{tab:data_provinces_MSA} and ~\ref{tab:data_provinces_DA} (Appendix A) show the distribution of the 100 province classes in our MSA and DA data splits, respectively. \textit{\textbf{Importantly, for all 4 subtasks, tweets in the TRAIN, DEV and TEST splits come from disjoint sets.} }

\subsection{Restrictions and Evaluation Metrics}
We follow the same general approach to managing the shared task as our first NADI in 2020. This includes providing participating teams with a set of restrictions that apply to all subtasks, and clear evaluation metrics. The purpose of our restrictions is to ensure fair comparisons and common experimental conditions. In addition, similar to NADI 2020, our data release strategy and our evaluation setup through the CodaLab online platform facilitated the competition management, enhanced timeliness of acquiring results upon system submission, and guaranteed ultimate transparency.\footnote{\url{https://codalab.org/}}

Once a team registered in the shared task, we directly provided the registering member with the data via a private download link. We provided the data in the form of the actual tweets posted to the Twitter platform, rather than tweet IDs. This guaranteed comparison between systems exploiting identical data. For all four subtasks, we provided clear instructions requiring participants not to use any external data. That is, teams were required to only use the data we provided to develop their systems and no other datasets regardless how these are acquired. For example, we requested that teams do not search nor depend on any additional user-level information such as geolocation. To alleviate these strict constraints and encourage creative use of diverse (machine learning) methods in system development, we provided an unlabeled dataset of 10M tweets in the form of tweet IDs. This dataset is in addition to our labeled TRAIN and DEV splits for the four subtasks. To facilitate acquisition of this unlabeled dataset, we also provided a simple script that can be used to collect the tweets. We encouraged participants to use these 10M unlabeled tweets in any way they wished.

For all four subtasks, the official metric is macro-averaged $F{_1}$ score obtained on blind test sets. We also report performance in terms of macro-averaged precision, macro-averaged recall and accuracy for systems submitted to each of the four subtasks. Each participating team was allowed to submit up to five runs for each subtask, and only the highest scoring run was kept as representing the team. Although official results are based only on a blind TEST set, we also asked participants to report their results on the DEV set in their papers. We setup four CodaLab competitions for scoring participant systems.\footnote{Links to the 
CodaLab competitions are as follows: Subtask~1.1: \url{https://competitions.codalab.org/competitions/27768}, Subtask~1.2: \url{https://competitions.codalab.org/competitions/27769}, Subtask~2.1: \url{https://competitions.codalab.org/competitions/27770}, Subtask~2.2: \url{https://competitions.codalab.org/competitions/27771}.
} We will keep the Codalab competition for each subtask live post competition, for researchers who would be interested in training models and evaluating their systems using the shared task TEST set. For this reason, we will not release labels for the TEST set of any of the subtasks.

\section{Shared Task Datasets}\label{sec:data}
We distributed two Twitter datasets, one in MSA and another in DA. Each tweet in each of these two datasets has two labels, one label for country level and another label for province level. For example, for the MSA dataset, the same tweet is assigned one out of 21 country labels (Subtask~1.1) and one out of 100 province labels (Subtask~2.1). The same applies to DA data, where each tweet is assigned a country label (Subtask~1.2) and a province label (Subtask~2.2). Similar to MSA, the tagset for DA data has 21 country labels and 100 province labels. In addition, as mentioned before, we made available an unlabeled dataset for optional use in any of the four subtasks. We now provide more details about both the labeled and unlabeled data.  

\subsection{Data Collection}

Similar to NADI 2020, we used the Twitter API to crawl data from 100 provinces belonging to 21 Arab countries for 10 months (Jan. to Oct., 2019).\footnote{Although we tried, we could not collect data from Comoros to cover all 22 Arab countries.} Next, we identified users who consistently and \textit{exclusively} tweeted from a single province during the whole 10 month period. We crawled up to 3,200 tweets from each of these users. We select only tweets assigned the Arabic language tag (\texttt{ar}) by Twitter. We lightly normalize tweets by removing usernames and hyperlinks, and add white space between emojis. Next, we remove retweets (i.e., we keep only tweets and replies). Then, we use character-level string matching to remove sequences that have $< 3$ Arabic tokens. 


\begin{figure}[h!]
  \centering
 \includegraphics[width=\linewidth]{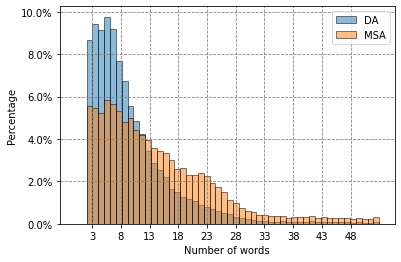}
  \caption{Distribution of tweet length (trimmed at $50$) in words in NADI-2021 labeled data.}
\end{figure}

Since the Twitter language tag can be wrong sometimes, we apply an effective in-house language identification tool on the tweets and replies to exclude any non-Arabic. This helps us remove posts in Farsi (\texttt{fa}) and Persian (\texttt{ps}) which Twitter wrongly assigned an Arabic language tag. Finally, to tease apart MSA from DA, we use the dialect-MSA model introduced in~\newcite{mageed2020marbert} (acc= $89.1$\%, F1= $88.6$\%).



\begin{table*}[!ht]
\centering
\begin{small}
\begin{tabular}{llr}
\toprule
\textbf{Country} & \textbf{Province} & \multicolumn{1}{c}{\textbf{Tweet}} \\\toprule

Algeria                              & Bouira                         & <شحال راكي تبيعي فيه... نجي ندي كونتيتي>                                                                                                                                                                                                                              \\\cdashline{2-3}

                                                         & Khenchela                      & <شحال يقلبو النم هاذوك لي يقدسو الراپرز ،!>                                                                                                                                                                                                                           \\ \cdashline{2-3}

                              & Oran                           & <راك زعفان مزية ربحتوا>                                                                                                                                                                                                                                               \\

                            \hline

    Egypt                          & Alexandria                     & <بس مش زي ما هنقدر نعدي حياتنا>                                                                                                                                                                                                                                       \\\cdashline{2-3}
                                & Minya                          & <بص انا كل مضميري هيبقي صاحي هجيلك تقتلهولي>                                                                                                                                                                                                                          \\\cdashline{2-3}


                              & Sohag                          &  <انا معنديش حد يفسحنى زي الوليه >\\

                              \hline

     KSA                   & Ar-Riyad                       & <و يعدين تجلسون تلعبون سماش !!>    \\\cdashline{2-3}



                         & Ash-Sharqiyah                  & <يابن الحلال اسمهم رجال فقراويه مايفهمون تبي حريمهم تفهم>                                                                                                                                                                                                             \\\cdashline{2-3}
                                                                                                              & Tabuk                          & <طيب ايش دخل الوزارة هذي وفاه طبيعيه>                                                                                                                                                                                                                                
\\ \hline
Morocco                              & Marrakech-Tensift-Al-Haouz     & <مافي ربح ولا كرش العصبية اشرب بنادول عشان ارتاح>                                                                                                                                                                                                                     \\\cdashline{2-3}
                              &Meknes-Tafilalet&<مراداون إنيخ أداخ إعفو ربي خ إمداكلن لحاجت !!!!>\\\cdashline{2-3}

                              & Souss-Massa-Draa               & <أييه نسيتهم ڭاع حتى هما متشيعييين بزاف>                                                                                     

                              \\ \hline

Oman                               
                                 & Ash-Sharqiyah                  & <يحصلج انتي اصلاً حد يعلق لج.>                                                                                    \\\cdashline{2-3}

                                 & Dhofar                         & <اسمعي قران ، ميبي تغفين عليه>                                                                                                                                                    \\\cdashline{2-3}

                                & Musandam                       & <ماني بقايل غلاتك هقوه وخابت بقول عين الحسود الله يجازيها>                                                                                                                                                                                                                                                                                                                                                                                                                                                    

                           \\ \hline
                            & Gaza-Strip & <احنا اليوم عاملين مفتول وانتو ؟؟>                  
\\\cdashline{2-3}
                 
Palestine & West-Bank  & <اتفقنا ع هيك>                                       \\\cdashline{2-3}
 & West-Bank  & <اليوم كنت فيهم بجننو>                      

\\ \hline

Sudan                                & Khartoum                       & \begin{tabular}[c]{@{}r@{}}<صرت عادي اشوف أشياء تقهرني واسكت.>\\<لأني ادري لو تكلمت م راح القي شي برضيني صرت انهي كلامي...>\end{tabular} \\\cdashline{2-3}

                                & Khartoum                       & \begin{tabular}[c]{@{}r@{}}<من وين جبتي كلامك دا..... إستندتي علي شنو >\\ <إنه لو قتلنا ح نكون قتلة...>\end{tabular}                                  \\\cdashline{2-3}                                                            
&Khartoum & <انا شكلي حأتبع سياسة قصي وابلك من طرف.>
\\ \hline

UAE                 & Abu-Dhabi                      & <يحسبني لاهِي عنه و أنا ملتهي به>                                                                                                                                                                                                                                     \\\cdashline{2-3}

                 & Dubai                          & <يـلي يبـا خـوتي بالـطيب خـاويته>                                                                                                                                                                                                                                     \\ \cdashline{2-3}

                 & Ras-Al-Khaymah                 & <احسك مني مليت>                                               <ل موسم نخربها بالاخر وما تدري شو المشكلة الأساسية> \\
\toprule
\end{tabular}%
\end{small}
\caption{Randomly picked DA tweets from select provinces and corresponding countries.}\label{tab:twt_exmpls_DA}
\end{table*}
\subsection{Data Sets}
To assign labels for the different subtasks, we use user {\it location} as a proxy for {\it  language variety labels} at both country and province levels. This applies to both our MSA and DA data. That is, we label tweets from each user with the country and province from which the user consistently posted for the \textit{whole} of the 10 months period. Although this method of label assignment is not ideal, it is still a reasonable approach for easing the bottleneck of data annotation. For both the MSA and DA data, across the two levels of classification (i.e., country and province), we randomly sample $21$K tweets for training (TRAIN), $5$K tweets for development (DEV), and $5$K tweets for testing (TEST). These three splits come from three disjoint sets of users. We distribute data for the four subtasks directly to participants in the form of actual tweet text. Table~\ref{tab:country} shows the distribution of tweets across the data splits over the 21 countries, for all subtasks. We provide the data distribution over the 100 provinces in Appendix A. More specifically, Table~\ref{tab:data_provinces_MSA} shows the province-level distribution of tweets for MSA (Subtask~2.1) and Table~\ref{tab:data_provinces_DA} shows the same for DA (Subtask~2.2). We provide examples DA tweets from a number of countries representing different regions in Table~\ref{tab:twt_exmpls_DA}. For each example in Table~\ref{tab:twt_exmpls_DA}, we list the province it comes from. Similarly, we provide example MSA data in Table~\ref{tab:twt_exmpls_MSA}.



\paragraph{Unlabeled 10M.} ~We shared 10M Arabic tweets with participants in the form of tweet IDs. We crawled these tweets in 2019. Arabic was identified using Twitter language tag (\texttt{ar}). This dataset does not have any labels and we call it UNLABELED 10M. We also included in our data package released to participants a simple script to crawl these tweets. Participants were free to use UNLABELED 10M for any of the four subtasks in any way they they see fits.\footnote{Datasets for all the subtasks and UNLABELED 10M are available at \url{https://github.com/UBC-NLP/nadi}. More information about the data format can be found in the accompanying README file.} We now present shared task teams and results.
\begin{table*}[h!]
\centering
\begin{small}
%
\end{small}
\caption{Randomly picked MSA tweets from select provinces and corresponding countries.}\label{tab:twt_exmpls_MSA}
\end{table*}
\begin{table*}[h!]

\centering
\small
\begin{tabular}{@{}llr@{}}
\toprule
\multicolumn{1}{c}{\textbf{Team}} & \multicolumn{1}{c}{\textbf{Affiliation}}    & \multicolumn{1}{c}{\textbf{Tasks}} \\ \midrule
\textbf{AraDial\_MJ}~\cite{althobaiti:2021:country-level}              & Taif Uni, KSA                        & 1.2                                \\
\textbf{Arizona}~\cite{issa:2021:country-level}                  & Uni of Arizona, USA                  & 1.2                                \\
\textbf{CairoSquad}~\cite{alkhamiss:2021:adapting}               & Microsoft, Egypt                            & all                 \\
\textbf{CS-UM6P}~\cite{mekki:2021:csum6p}                  & Mohammed VI Polytech, Morocco & all                 \\
\textbf{NAYEL}~\cite{nayel:2021:data-driven}                    & Benha Uni, Egypt                     & all                 \\
\textbf{Phonemer}~\cite{wadhawan:2021:daialect}                 & Flipkart Private Limited, India             & all                 \\
\textbf{Speech Trans}~\cite{lichouri:2021:arabic-dialect}       & CRSTDLA, Algeria            & 1.1, 1.2                           \\ \bottomrule
\end{tabular}
\caption{List of teams that participated in one or more of the four subtasks \textit{and} submitted a system description paper.}~\label{tab:teams}
\end{table*}
\begin{table*}[ht!]
\centering
\small
\begin{tabular}{l
      S[table-format=2.2 (2)]
      S[table-format=2.2 (2)]
      S[table-format=2.2 (2)]
      S[table-format=2.2 (2)]}
\toprule
\textbf{Team}               & \multicolumn{1}{c}{\textbf{$F_1$}} & \multicolumn{1}{c}{\textbf{Acc}} & \multicolumn{1}{c}{\textbf{Precision}} & \multicolumn{1}{c}{\textbf{Recall}} \\ \midrule
\textbf{CairoSquad}         & 22.38 (1)                       & 35.72 (1)                             & 31.56 (1)                              & 20.66 (1)                           \\
\textbf{Phonemer}           & 21.79 (2)                       & 32.46 (3)                             & 30.03 (3)                              & 19.95 (2)                           \\
\textbf{CS-UM6P}            & 21.48 (3)                       & 33.74 (2)                             & 30.72 (2)                              & 19.70 (3)                           \\
\textbf{Speech Translation} & 14.87 (4)                       & 24.32 (4)                             & 18.95 (4)                              & 13.85 (4)                           \\
\textbf{\color{blue}Our Baseline II}     & 14.15                           & 24.76                                 & 20.01                                  & 13.21                               \\
\textbf{NAYEL}              & 12.99 (5)                       & 23.24 (5)                             & 15.09 (5)                              & 12.46 (5)                           \\
\textbf{\color{blue}Our Baseline I}    & 1.57                            & 19.78                                 & 0.94                                   & 4.76                                \\ \bottomrule
\end{tabular}
\caption{Results for Subtask 1.1 (country-level MSA). The numbers in parentheses are the ranks. The table is sorted on the $macro-F$\textsubscript{1} score, the official metric. }\label{tab:sub11_res}
\end{table*}
\begin{table*}[h!]
\centering
\small
\begin{tabular}{l
      S[table-format=2.2 (2)]
      S[table-format=2.2 (2)]
      S[table-format=2.2 (2)]
      S[table-format=2.2 (2)]}
\toprule
\multicolumn{1}{c}{\textbf{Team}} & \multicolumn{1}{c}{\textbf{$F_1$}} & \multicolumn{1}{c}{\textbf{Acc}} & \multicolumn{1}{c}{\textbf{Precision}} & \multicolumn{1}{c}{\textbf{Recall}} \\ \midrule
\textbf{CairoSquad}               & 32.26 (1)                       & 51.66 (1)                             & 36.03 (1)                              & 31.09 (1)                           \\
\textbf{CS-UM6P}                  & 30.64 (2)                       & 49.50 (2)                             & 32.91 (2)                              & 30.34 (2)                           \\
\textbf{IDC team}                 & 26.10 (3)                       & 42.70 (4)                             & 27.04 (4)                              & 25.88 (3)                           \\
\textbf{Phonemer}                 & 24.29 (4)                       & 44.14 (3)                             & 30.24 (3)                              & 23.70 (4)                           \\
\textbf{Speech Translation}       & 21.49 (5)                       & 40.54 (5)                             & 26.75 (5)                              & 20.36 (6)                           \\
\textbf{Arizona}                  & 21.37 (6)                       & 40.46 (6)                             & 26.32 (6)                              & 20.78 (5)                           \\
\textbf{AraDial\_MJ}              & 18.94 (7)                       & 35.94 (8)                             & 21.58 (8)                              & 18.28 (7)                           \\
\bottomrule
\end{tabular}
\caption{Results for Subtask 1.2 (country-level DA)}\label{tab:sub12_res}
\end{table*}
\begin{table*}[h!]
\centering
\small
\begin{tabular}{l
      S[table-format=1.2 (2)]
      S[table-format=1.2 (2)]
      S[table-format=1.2 (2)]
      S[table-format=1.2 (2)]}
\toprule
\multicolumn{1}{c}{\textbf{Team}} & \multicolumn{1}{c}{\textbf{$F_1$}} & \multicolumn{1}{c}{\textbf{Acc}} & \multicolumn{1}{c}{\textbf{Precision}} & \multicolumn{1}{c}{\textbf{Recall}} \\ \midrule
\textbf{CairoSquad}               & 6.43 (1)                        & 6.66 (1)                              & 7.11 (1)                               & 6.71 (1)                            \\
\textbf{Phonemer}                 & 5.49 (2)                        & 6.00 (2)                              & 6.17 (2)                               & 6.07 (2)                            \\
\textbf{CS-UM6P}                  & 5.35 (3)                        & 5.72 (3)                              & 5.71 (3)                               & 5.75 (3)                            \\
\textbf{NAYEL}                    & 3.51 (4)                        & 3.38 (4)                              & 4.09 (4)                               & 3.45 (4)                            \\
\textbf{\color{blue}Our Baseline II}           & 3.39                            & 3.48                                  & 3.68                                   & 3.49                                \\
\textbf{\color{blue}Our Baseline I}          & 0.02                            & 1.02                                  & 0.01                                   & 1.00                                \\ \bottomrule
\end{tabular}
\caption{Results for Subtask 2.1 (province-level MSA).}\label{tab:sub21_res}
\end{table*}
\begin{table*}[h!]
\centering
\small
\begin{tabular}{l
      S[table-format=1.2 (2)]
      S[table-format=1.2 (2)]
      S[table-format=1.2 (2)]
      S[table-format=1.2 (2)]}
\toprule
\multicolumn{1}{c}{\textbf{Team}} & \multicolumn{1}{c}{\textbf{$F_1$}} & \multicolumn{1}{c}{\textbf{Acc}} & \multicolumn{1}{c}{\textbf{Precision}} & \multicolumn{1}{c}{\textbf{Recall}} \\ \midrule
\textbf{CairoSquad}               & 8.60 (1)                        & 9.46 (1)                              & 9.07 (1)                               & 9.33 (1)                            \\
\textbf{CS-UM6P}                  & 7.32 (2)                        & 7.92 (2)                              & 7.73 (2)                               & 7.95 (2)                            \\
\textbf{NAYEL}                    & 4.55 (3)                        & 4.80 (3)                              & 4.71 (3)                               & 4.55 (4)                            \\
\textbf{Phonemer}                 & 4.37 (4)                        & 5.32 (4)                              & 4.49 (4)                               & 5.19 (3)                            \\
\textbf{\color{blue}Our Baseline II}           & 4.08                            & 4.18                                  & 4.54                                   & 4.22                                \\
\textbf{\color{blue}Our Baseline I}          & 0.02                            & 1.06                                  & 0.01                                   & 1.00                                \\ \bottomrule
\end{tabular}
\caption{Results for Subtask 2.2 (province-level DA).}\label{tab:sub22_res}
\end{table*}
\begin{table*}[h!]
\centering
\small  
\begin{tabular}{@{}lHS[table-format=2.2]cccccccccccc@{}}
\toprule
\multicolumn{1}{c}{\textbf{}}     & \textbf{}          & \multicolumn{1}{c}{\textbf{}}   & \multicolumn{6}{c}{\textbf{Features}}                                                                              & \multicolumn{6}{c}{\textbf{Techniques}}                                                                                                 \\ \cmidrule(lr){4-9} \cmidrule(lr){10-15}
\multicolumn{1}{c}{\textbf{Team}} & \textbf{\rotatebox[origin=c]{90}{\# submit}} & \multicolumn{1}{c}{\rotatebox[origin=c]{90}{$\bf F_1$}} & \textbf{\rotatebox[origin=c]{90}{N-gram}} & \textbf{\rotatebox[origin=c]{90}{TF-IDF}} & \textbf{\rotatebox[origin=c]{90}{Linguistics}} & \textbf{\rotatebox[origin=c]{90}{Word embeds}} & \textbf{\rotatebox[origin=c]{90}{PMI}} & \textbf{\rotatebox[origin=c]{90}{Sampling}} & \textbf{\rotatebox[origin=c]{90}{Classical ML}} & \textbf{\rotatebox[origin=c]{90}{Neural nets}} & \textbf{\rotatebox[origin=c]{90}{Transformer}} & \textbf{\rotatebox[origin=c]{90}{Ensemble}} & \textbf{\rotatebox[origin=c]{90}{Multitask}} & \textbf{\rotatebox[origin=c]{90}{Semi-super}} \\ \midrule
\multicolumn{15}{c}{\textbf{\colorbox{orange!15}{SUBTASK 1.1}}}                                                                                                                                                                                                                                                                                                                                                                  \\ \midrule
CairoSquad                                         & 2                                   & 22.38                                            &                 &                 &                      &                      &              &                   &                       &                      & \checkmark                  & \checkmark               &                    &                          \\
\bottomrule
\end{tabular}
\caption{Summary of approaches used by participating teams. PMI: poinwise mutual information. Classical ML refers to any non-neural machine learning methods such as naive Bayes and support vector machines. The term ``neural nets" refers to any model based on neural networks (e.g., FFNN, RNN, and CNN) except Transformer models. Transformer refers to neural networks based on a Transformer architecture such as BERT. The table is sorted by official metric, $macro-F$\textsubscript{1}. We only list teams that submitted a description paper. ``Semi-super" indicates that the model is trained with semi-supervised learning.}\label{tab:system_sum}
\end{table*}

\section{Shared Task Teams \& Results}\label{sec:teams_res}

\subsection{Our Baseline Systems}
We provide two simple baselines, Baseline I and Baseline II, for each of the four subtasks. \textbf{Baseline~I} is based on the majority class in the TRAIN data for each subtask. It performs at  \textit{$F_1=1.57\%$} and \textit{$accuracy=19.78\%$} for Subtask 1.1, \textit{$F_1=1.65\%$} and \textit{$accuracy=21.02\%$} for Subtask 1.2,  \textit{$F_1=0.02\%$} and \textit{$accuracy=1.02\%$} for Subtask 2.1, and  \textit{$F_1=0.02\%$} and \textit{$accuracy=1.06\%$}  for Subtask 2.2. 

\textbf{Baseline~II} is a fine-tuned multi-lingual BERT-Base model (mBERT)\footnote{https://github.com/google-research/bert}. More specifically, we fine-tune mBERT for $20$ epochs with a learning rate of $2e-5$, and batch size of $32$. The maximum length of input sequence is set as $64$ tokens. We evaluate the model at the end of each epoch and choose the best model on our DEV set. We then report the best model on the TEST set. Our best mBERT model obtains \textit{$F_1=14.15\%$} and \textit{$accuracy=24.76\%$} on Subtask 1.1, \textit{$F_1=18.02\%$} and \textit{$accuracy=33.04\%$} on Subtask 1.2, \textit{$F_1=3.39\%$} and \textit{$accuracy=3.48\%$} on Subtask 2.1, and \textit{$F_1=4.08\%$} and \textit{$accuracy=4.18\%$} on Subtask 2.2 as Tables~\ref{tab:sub11_res},~\ref{tab:sub12_res},~\ref{tab:sub21_res}, and~\ref{tab:sub22_res}, respectively. 



\subsection{Participating Teams}

We received a total of $53$ unique team registrations. After evaluation phase, we received a total of $68$ submissions. The breakdown across the subtasks is as follows: $16$ submissions for Subtask 1.1 from five teams, $27$ submissions for Subtask 1.2 from eight teams, $12$ submissions for Subtask 2.1 from four teams, and $13$ submissions for Subtask 2.2 from four teams. Of participating teams, seven teams submitted description papers, all of which we accepted for publication. Table~\ref{tab:teams} lists the seven teams.

\subsection{Shared Task Results}
Table~\ref{tab:sub11_res} presents the best TEST results for all $5$ teams who submitted systems for Subtask 1.1. Based on the official metric, $macro-F_1$, CairoSquad obtained the best performance with $22.38\%$ $F_1$ score. Table~\ref{tab:sub12_res} presents the best TEST results of each of the eight teams who submitted systems to Subtask 1.2. Team CairoSquad achieved the best $F_1$ score that is $32.26\%$. Table~\ref{tab:sub21_res} shows the best TEST results for all four teams who submitted systems for Subtask 2.1. CairoSquad achieved the best performance with $6.43\%$ $F_1$ score. 

Table~\ref{tab:sub22_res} provides the best TEST results of each of the four teams who submitted systems to Subtask 2.2. CairoSquad also achieved the best performance with $8.60\%$.\footnote{The full sets of results for Subtask 1.1, 1.2, 2.1, and 2.2 are in Tables~\ref{tab:sub11_res_full}, \ref{tab:sub12_res_full}, \ref{tab:sub21_res_full} and ~\ref{tab:sub21_res_full}, respectively, in Appendix A.} 

\def\checkmark{\tikz\fill[scale=0.4](0,.35) -- (.25,0) -- (1,.7) -- (.25,.15) -- cycle;} 


\subsection{General Description of Submitted Systems}

In Table~\ref{tab:system_sum}, we provide a high-level description of the systems submitted to each subtask. For each team, we list their best score of each subtask, the features employed, and the methods adopted/developed. As can be seen from the table, the majority of the top teams have used Transformers. Specifically, team CairoSquad and CS-UM6P developed their system utilizing MARBERT~\cite{mageed2020marbert}, a pre-trained Transformer language model tailored to Arabic dialects and the domain of social media. Team Phonemer utilized AraBERT~\cite{antoun2020arabert} and AraELECTRA~\cite{antoun2020araelectra}. Team CairSquad apply adapter modules~\cite{houlsby2019parameter} and vertical attention to MARBERT fine-tuning. CS-UM6P fine-tuned MARBERT on country-level and province-level jointly by multi-task learning. The rest of participating teams have either used a type of neural networks other than Transformers or resorted to linear machine learning models, usually with some form of ensembling.   

\section{Conclusion and Future Work}\label{sec:conc}
We presented the findings and results of the NADI 2021 shared task. We described our datasets across the four subtasks and the logistics of running the shared task. We also provided a panoramic description of the methods used by all participating teams. The results show that distinguishing the language variety of short texts based on small geographical regions of origin is possible, yet challenging. The total number of submissions during official evaluation (n=$68$ submissions from $8$ unique teams), as well as the number of teams who registered and acquired our datasets (n=$53$ unique teams) reflects a continued interest in the community and calls for further work in this area.
In the future, we plan to host a third iteration of the NADI shared task that will use new datasets and encourage novel solutions to the set of problems introduced in NADI 2021. As results show all the fours subtasks remain challenging.

\section*{Acknowledgments}
We gratefully acknowledge the support of the Natural Sciences and Engineering Research Council of Canada, the Social Sciences Research Council of Canada, Compute Canada, and UBC Sockeye.

\normalem
\bibliography{dlnlp,camel-bib-v2,NADI2021-paper,extra}
\bibliographystyle{acl_natbib}

\appendix
\appendixpage
\addappheadtotoc

\section{Data}\label{sec:appendix_data}
We provide the distribution Distribution of the NADI 2021 MSA data over provinces, by country (Subtask~2.1), across our our data splits in Table~\ref{tab:data_provinces_MSA}. Similarly, Table~\ref{tab:data_provinces_DA} shows the distribution of the DA data over provinces for all countries (Subtask~2.2) in our data splits.

%

\begin{table*}[ht]
\setlength{\tabcolsep}{2pt}
\centering
\small
\begin{tabular}{lccc|lccc}
\hline
\textbf{Province   Name} & \textbf{TRAIN} & \textbf{DEV} & \textbf{TEST} & \textbf{Province Name}         & \textbf{TRAIN} & \textbf{DEV} & \textbf{TEST} \\ \hline
ae\_Abu-Dhabi       & 211   & 51  & 51   & kw\_Jahra                      & 211   & 52  & 51   \\ 
ae\_Dubai           & 211   & 52  & 51   & lb\_Akkar                      & 211   & 52  & 39   \\ 
ae\_Ras-Al-Khaymah  & 211   & 51  & 51   & lb\_North-Lebanon              & 211   & 51  & 51   \\ 
bh\_Capital         & 211   & 51  & 51   & lb\_South-Lebanon              & 211   & 52  & 51   \\ 
dj\_Djibouti        & 211   & 52  & 51   & ly\_Al-Butnan                  & 211   & 52  & 51   \\ 
dz\_Batna           & 211   & 52  & 51   & ly\_Al-Jabal-al-Akhdar         & 211   & 52  & 52   \\ 
dz\_Biskra          & 211   & 52  & 51   & ly\_Benghazi                   & 211   & 51  & 51   \\
\hline
\end{tabular}%
\caption{Distribution of the NADI 2021 MSA data over provinces, by country, across our TRAIN, DEV, and TEST splits (Subtask~2.1).}\label{tab:data_provinces_MSA} 
\end{table*}

\begin{table*}[h!]
\centering
\small
\begin{tabular}{lccc|lccc}
\hline
\textbf{Province   Name} & \textbf{TRAIN} & \textbf{DEV} & \textbf{TEST} & \textbf{Province Name}         & \textbf{TRAIN} & \textbf{DEV} & \textbf{TEST} \\ \hline
ae\_Abu-Dhabi            & 214            & 52           & 52            & kw\_Jahra                      & 215            & 53           & 53            \\ 
ae\_Dubai                & 214            & 53           & 53            & lb\_Akkar                      & 215            & 53           & 14            \\ 
ae\_Ras-Al-Khaymah       & 214            & 52           & 53            & lb\_North-Lebanon              & 215            & 52           & 53            \\ 
bh\_Capital              & 215            & 52           & 52            & lb\_South-Lebanon              & 214            & 52           & 53            \\ 
dj\_Djibouti             & 215            & 27           & 7             & ly\_Al-Butnan                  & 214            & 52           & 53            \\ 
dz\_Batna                & 215            & 34           & 10            & ly\_Al-Jabal-al-Akhdar         & 215            & 53           & 53            \\ 
dz\_Biskra               & 215            & 53           & 53            & ly\_Benghazi                   & 214            & 52           & 52            \\
\hline
\end{tabular}%
\caption{Distribution of the NADI 2021 DA data over provinces, by country, across our TRAIN, DEV, and TEST splits (Subtask~2.2).}\label{tab:data_provinces_DA} 
\end{table*}

\section{Shared Task Teams \& Results}\label{sec:appendix_results}
We provide full results for all four subtasks. Table~\ref{tab:sub11_res_full} shows full results for Subtask~1.1, Table~\ref{tab:sub12_res_full} for Subtask~1.2, Table~\ref{tab:sub21_res_full} for Subtask~2.1, and Table~\ref{tab:sub22_res_full} for Subtask~2.2. 
\begin{table*}[h!]
\centering
\small
\begin{tabular}{l
      S[table-format=2.2 (2)]
      S[table-format=2.2 (2)]
      S[table-format=2.2 (2)]
      S[table-format=2.2 (2)]}
\toprule
\multicolumn{1}{c}{\textbf{Team}} & \multicolumn{1}{c}{\textbf{$F_1$}} & \multicolumn{1}{c}{\textbf{Acc}} & \multicolumn{1}{c}{\textbf{Precision}} & \multicolumn{1}{c}{\textbf{Recall}} \\ \midrule
CairoSquad                        & 22.38 (1)                       & 35.72 (1)                             & 31.56 (3)                              & 20.66 (1)                           \\
CairoSquad                        & 21.97 (2)                       & 34.90 (2)                             & 30.01 (7)                              & 20.15 (2)                           \\
Phonemer                          & 21.79 (3)                       & 32.46 (6)                             & 30.03 (6)                              & 19.95 (4)                           \\
Phonemer                          & 21.66 (4)                       & 31.70 (7)                             & 28.46 (8)                              & 20.01 (3)                           \\
CS-UM6P                           & 21.48 (5)                       & 33.74 (4)                             & 30.72 (5)                              & 19.70 (5)                           \\
CS-UM6P                           & 20.91 (6)                       & 33.84 (3)                             & 31.16 (4)                              & 19.09 (6)                           \\
Phonemer                          & 20.78 (7)                       & 32.96 (5)                             & 37.69 (1)                              & 18.42 (8)                           \\
\bottomrule
\end{tabular}
\caption{Full results for Subtask 1.1 (country-level MSA). The numbers in parentheses are the ranks. The table is sorted on the  $macro-F$\textsubscript{1} score, the official metric.}\label{tab:sub11_res_full} 
\end{table*}
\begin{table*}[ht]
\small
\centering
\begin{tabular}{l
      S[table-format=2.2 (2)]
      S[table-format=2.2 (2)]
      S[table-format=2.2 (2)]
      S[table-format=2.2 (2)]}
\toprule
\multicolumn{1}{c}{\textbf{Team}} & \multicolumn{1}{c}{\textbf{$F_1$}} & \multicolumn{1}{c}{\textbf{Acc}} & \multicolumn{1}{c}{\textbf{Precision}} & \multicolumn{1}{c}{\textbf{Recall}} \\ \midrule
CairoSquad                        & 32.26 (1)                       & 51.66 (1)                             & 36.03 (1)                              & 31.09 (1)                           \\
CairoSquad                        & 31.04 (2)                       & 51.02 (2)                             & 35.01 (2)                              & 30.62 (2)                           \\
CS-UM6P                           & 30.64 (3)                       & 49.50 (4)                             & 32.91 (6)                              & 30.34 (3)                           \\
CS-UM6P                           & 30.14 (4)                       & 48.94 (5)                             & 33.20 (4)                              & 30.21 (4)                           \\
CS-UM6P                           & 29.08 (5)                       & 50.30 (3)                             & 34.99 (3)                              & 29.04 (5)                           \\
IDC team                          & 26.10 (6)                       & 42.70 (9)                             & 27.04 (11)                             & 25.88 (6)                           \\
Phonemer                          & 24.29 (7)                       & 44.14 (6)                             & 30.24 (7)                              & 23.70 (7)                           \\
\bottomrule
\end{tabular}
\caption{Full results for Subtask 1.2 (country-level DA).}\label{tab:sub12_res_full}
\end{table*}
\begin{table*}[ht]
\small
\centering
\begin{tabular}{l
      S[table-format=1.2 (2)]
      S[table-format=1.2 (2)]
      S[table-format=1.2 (2)]
      S[table-format=1.2 (2)]}
\toprule
\multicolumn{1}{c}{\textbf{Team}} & \multicolumn{1}{c}{\textbf{$F_1$}} & \multicolumn{1}{c}{\textbf{Acc}} & \multicolumn{1}{c}{\textbf{Precision}} & \multicolumn{1}{c}{\textbf{Recall}} \\ \midrule
CairoSquad                        & 6.43 (1)                        & 6.66 (1)                              & 7.11 (1)                               & 6.71 (1)                            \\
CairoSquad                        & 5.81 (2)                        & 6.24 (2)                              & 6.26 (2)                               & 6.33 (2)                            \\
Phonemer                          & 5.49 (3)                        & 6.00 (3)                              & 6.17 (3)                               & 6.07 (3)                            \\
Phonemer                          & 5.43 (4)                        & 5.96 (4)                              & 6.12 (4)                               & 6.02 (4)                            \\
CS-UM6P                           & 5.35 (5)                        & 5.72 (6)                              & 5.71 (7)                               & 5.75 (6)                            \\
Phonemer                          & 5.30 (6)                        & 5.84 (5)                              & 5.97 (6)                               & 5.90 (5)                            \\
CS-UM6P                           & 5.12 (7)                        & 5.50 (7)                              & 5.24 (8)                               & 5.53 (7)                            \\
\bottomrule
\end{tabular}
\caption{Full results for Subtask 2.1 (province-level MSA).}\label{tab:sub21_res_full} 
\end{table*}
\begin{table*}[ht]
\small
\centering
\begin{tabular}{l
      S[table-format=1.2 (2)]
      S[table-format=1.2 (2)]
      S[table-format=1.2 (2)]
      S[table-format=1.2 (2)]}
\toprule
\multicolumn{1}{c}{\textbf{Team}} & \multicolumn{1}{c}{\textbf{$F_1$}} & \multicolumn{1}{c}{\textbf{Acc}} & \multicolumn{1}{c}{\textbf{Precision}} & \multicolumn{1}{c}{\textbf{Recall}} \\ \midrule
CairoSquad                        & 8.60 (1)                        & 9.46 (1)                              & 9.07 (1)                               & 9.33 (1)                            \\
CairoSquad                        & 7.88 (2)                        & 8.78 (2)                              & 8.27 (2)                               & 8.66 (2)                            \\
CS-UM6P                           & 7.32 (3)                        & 7.92 (4)                              & 7.73 (4)                               & 7.95 (3)                            \\
CS-UM6P                           & 7.29 (4)                        & 8.04 (3)                              & 8.17 (3)                               & 7.90 (4)                            \\
CS-UM6P                           & 5.30 (5)                        & 6.90 (5)                              & 7.00 (5)                               & 6.82 (5)                            \\
NAYEL                             & 4.55 (6)                        & 4.80 (10)                             & 4.71 (6)                               & 4.55 (10)                           \\
NAYEL                             & 4.43 (7)                        & 4.88 (9)                              & 4.59 (8)                               & 4.62 (9)                            \\
\bottomrule
\end{tabular}
\caption{Full results for Subtask 2.2 (province-level DA).}\label{tab:sub22_res_full}
\end{table*}

\end{document}